\documentclass[10pt,journal,compsoc]{IEEEtran}

\usepackage{hyperref}       
\usepackage{url}            
\usepackage{booktabs}       
\usepackage{amsfonts}       
\usepackage{nicefrac}       
\usepackage{microtype}      %

\usepackage{color}
\usepackage{wrapfig}
\usepackage{graphicx}
\usepackage{amsmath}
\usepackage{enumitem}
\usepackage{threeparttable}
\usepackage{multirow}
\usepackage{xcolor}
\usepackage{adjustbox}
\usepackage{footnote}

\usepackage{mathrsfs}
\usepackage{array}
\usepackage{marvosym}

\usepackage{epsfig}
\usepackage{amssymb}
\usepackage{float}
\usepackage{bm}
\usepackage{caption}
\usepackage{ulem}

\ifCLASSOPTIONcompsoc
  \usepackage[nocompress]{cite}
\else
  \usepackage{cite}
\fi

\ifCLASSINFOpdf
\else
\fi

\begin{document}

\title{Syntax Customized Video Captioning by Imitating Exemplar Sentences}

\author{Yitian Yuan, Lin Ma, and Wenwu Zhu,~\IEEEmembership{Fellow,~IEEE}
\IEEEcompsocitemizethanks{

\IEEEcompsocthanksitem Y. Yuan is with Meituan, Beijing, China. This work was done when Yitian Yuan was at Tsinghua University, Beijing, China. E-mail: yuanyitian@foxmail.com. 

\IEEEcompsocthanksitem L. Ma is with Meituan, Beijing, China. E-mail: forest.linma@gmail.com. 

\IEEEcompsocthanksitem W. Zhu is with the Department of Computer Science and Technology, Tsinghua University, Beijing, China. W. Zhu is the corresponding author. E-mail: wwzhu@tsinghua.edu.cn.
}

\thanks{Manuscript received XXXX; revised XXXX.}}

\markboth{IEEE TRANSACTIONS ON PATTERN ANALYSIS AND MACHINE INTELLIGENCE,~Vol.~x, No.~x, March~2021}%
{Yuan \MakeLowercase{\textit{et al.}}:
Syntax Customized Video Captioning by Imitating Exemplar Sentences}

\IEEEtitleabstractindextext{%
\begin{abstract}
Enhancing the diversity of sentences to describe video contents is an important problem arising in recent video captioning research. In this paper, we explore this problem from a novel perspective of customizing video captions by imitating exemplar sentence syntaxes. Specifically, given a video and any syntax-valid exemplar sentence, we introduce a new task of  Syntax Customized Video Captioning (SCVC) aiming to generate one caption which not only semantically describes the video contents but also syntactically imitates the given exemplar sentence. To tackle the SCVC task, we propose a novel video captioning model, where a hierarchical sentence syntax encoder is firstly designed to extract the syntactic structure of the exemplar sentence, then a syntax conditioned caption decoder is devised to generate the syntactically structured caption expressing video semantics. As there is no available syntax customized groundtruth video captions, we tackle such a challenge by proposing a new training strategy, which leverages the traditional pairwise video captioning data and our collected exemplar sentences to accomplish the model learning. Extensive experiments, in terms of semantic, syntactic, fluency, and diversity evaluations, clearly demonstrate our model capability to generate syntax-varied and semantics-coherent video captions that well imitate different exemplar sentences with enriched diversities. Code is available at \url{{https://github.com/yytzsy/Syntax-Customized-Video-Captioning}}.
\end{abstract}

}

\maketitle

\IEEEdisplaynontitleabstractindextext

\IEEEpeerreviewmaketitle

\IEEEraisesectionheading{
\section{Introduction}
\label{sec:introduction}}

\begin{figure*}
\setlength{\abovecaptionskip}{0.0in}
\setlength{\belowcaptionskip}{0.0in}
\includegraphics[width=1.0\textwidth]{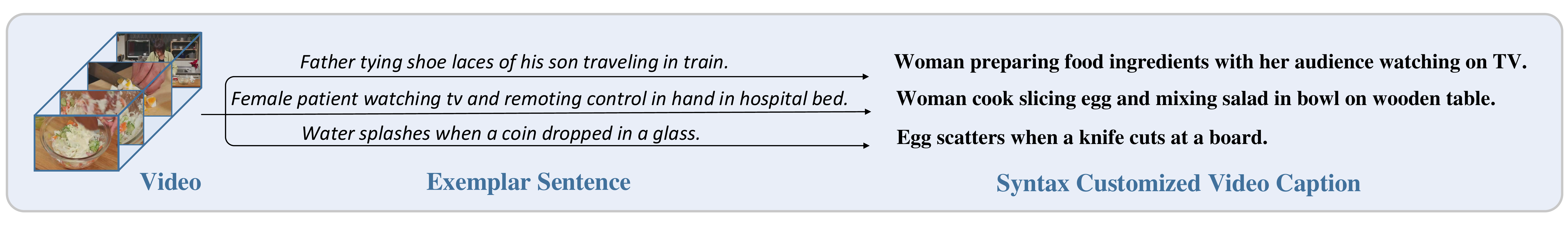} 
\captionof{figure}{The syntax customized video captioning task by imitating different exemplar sentences. It can be observed that the generated syntax customized video captions by our model can not only precisely depict the video contents, but also highly resemble the syntactic structures of the given exemplar sentences.}
\label{fig:task_definition}
\end{figure*}

\IEEEPARstart{V}ideo captioning \cite{chen2019deep,rohrbach2013translating,venugopalan2015sequence}, which aims to automatically generate natural language sentences to describe video contents, has aroused great interest recent years. Inspired by the success of the sequence-to-sequence model \cite{sutskever2014sequence} in neural machine translation, the encoder-decoder architecture \cite{venugopalan2014translating} leveraging CNNs and RNNs to encode video semantics and utilizing RNNs to decode sentences has become a common and effective configuration for most video captioning approaches.

Previous methods \cite{baraldi2017hierarchical,pan2016jointly,venugopalan2015sequence,wang2018reconstruction,xu2017learning,yao2015describing} are mostly trained with the (conditional) maximum likelihood objective, which encourages the use of the $n$-grams that appeared in the training samples. Consequently, the generated sentences will bear high resemblance to training sentences in detailed wording~\cite{dai2017towards}. For example, the generated sentences by the captioning models trained on MSRVTT~\cite{xu2016msr} often present similar syntactic structures like ``\texttt{A is doing B at/on/when C}". Although such generated captions can describe the video contents, the monotonous and plain sentence forms are of very limited linguistic diversity and expression variability.
Compared with those captioning models, people can express things more freely and vividly in daily life. Such an ability, on the one hand, is due to that we have the basic syntactic knowledge about sentence organization. On the other hand, we can acquire abundant corpora during daily reading and talking, and learn how to express things. Therefore, is it possible to learn and leverage sentence syntactic information to guide the video caption generation? Or, more intuitively, can we use the human-edited sentences to customize video captions so as to strengthen the captioning variability in expression?

To answer the above questions, we propose a novel task, namely Syntax Customized Video Captioning (SCVC) in this paper. As shown in Figure \ref{fig:task_definition}, given a video and any syntax-valid exemplar sentence, we aim to generate one caption, which should not only express the video semantics but also follow the syntactic structure of the given exemplar sentence. Since exemplar sentences are easy to acquire with no restrictions, our proposed task provides a promising way to strengthen the diversity and expressiveness of video captioning.

In order to solve the SCVC task, our proposed video captioning model consists of three fully coupled components. Firstly, one hierarchical sentence syntax encoder is proposed to capture the syntactic information of the exemplar sentence, in specific a character-level LSTM and a word-level LSTM are stacked to characterize the local lexical features and the global syntactic structures, respectively. Then, one video semantic encoder is introduced to represent the video semantics. Finally, we design a syntax conditioned caption decoder realized in a two-layer LSTM, in which one layer relies on the exemplar syntactic information to modulate the LSTM for customizing the caption syntax, and the subsequent layer relies on the video features to modulate the LSTM for generating the syntactically structured caption describing video semantics. As we do not have the syntax customized groundtruth captions, we propose a new training strategy, which fully leverages the syntactic and semantic information residing in both conventional video captioning data and our collected auxiliary exemplar sentences to train the overall model, thus enabling the generation of video captions with customized syntaxes.

Our main contributions are summarized as follows.
\begin{itemize}
    \item A novel Syntax Customized Video Captioning (SCVC) task is proposed to enrich the diversity and expressiveness of video descriptions.
    \item A novel video captioning model is designed, with a hierarchical sentence syntax encoder and a syntax conditioned caption decoder effectively encoding the exemplar sentence syntaxes and controlling the syntactic structures of the generated video captions, respectively.
    \item A new training strategy is proposed to fully utilize both the public video captioning data and the collected exemplar sentences to train our overall model. Extensive experiments also verify our model capability to generate various syntax customized video captions with enriched diversities.
\end{itemize}

\section{Related Works}

Video captioning has been extensively studied in the past few years. From pioneering template-based methods \cite{guadarrama2013youtube2text,rohrbach2014coherent,rohrbach2013translating,xu2015jointly} which defined special grammar rules to compose captions, to recent sequence-to-sequence architectures which leveraged RNNs to encode videos and then decode captions sequentially \cite{baraldi2017hierarchical,pan2016jointly,pasunuru2017reinforced,venugopalan2015sequence,wang2018reconstruction,xu2017learning,yao2015describing}, numerous improvements have been achieved. However, most of these previous approaches are trained to select words with the maximum probability sampled from the learned distribution of the training corpus, resulting in a monotonous set of generated captions bearing high resemblance to the training data.

The above works mainly aim at generating precise captions to describe visual contents for keeping fidelity, inspired by the research of text style transfer~\cite{fu2018style,jin2020deep,wu2020mask,dos2018fighting,cheng2020contextual} in the natural language processing community, some recent works for video captioning further put more emphasis on improving the expressiveness or diversity of the predicted video descriptive sentences \cite{chen2020say,dai2017towards,deshpande2019fast,duan2018weakly,Gan2017StyleNet,Mathews2018SemStyle,wang2019controllable,xiao2019diverse,YouImage}. Specifically, Gan \textit{et al.} proposed a StyleNet to control the style in the captioning process so as to produce attractive captions with the desired style like romantic or humorous \cite{Gan2017StyleNet}. You \textit{et al.} supplied the sentiment label as one additional dimension of the input feature, so as to make the model learn to control the injection of sentiment words in captions for sentiment-conveying \cite{YouImage}. Wang \textit{et al.} proposed to sample a Part-of-Speech (POS) sequence based on the video representation. By manually altering the sampled POS, the subsequently predicted captions can also be modified \cite{wang2019controllable}. Chen \textit{et al.} proposed fine-grained control of image caption generation with an abstract scene graph, which makes the level of details such as attributes or relationships can be captured in the caption generation process \cite{chen2020say}.

Compared with using the style label/sentiment label/POS tag/scene graph to control caption generation, directly leveraging the exemplar sentences to customize the generated captions is  more intuitive and natural. Moreover, since the exemplar sentences are easy to obtain and of various syntactic structures, the expressiveness and diversity of video captions can thereby be enriched through imitating those syntax-varied exemplars.

\begin{figure*}
\centering
\includegraphics[width=1.0\textwidth]{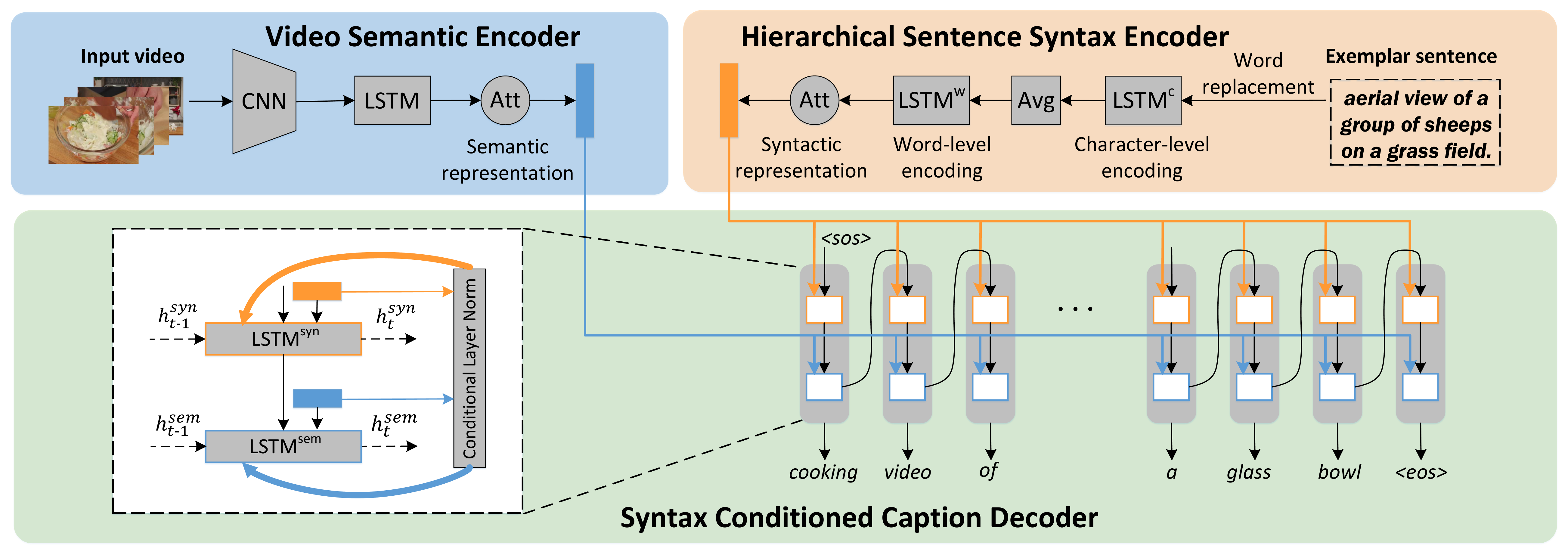} 
\caption{ Framework of our proposed model for the syntax customized video captioning task, which consists of three fully coupled components, \textit{i.e.}, a hierarchical sentence syntax encoder extracting the syntactic structure of the exemplar sentence, a video semantic encoder yielding the video semantic representation, and a syntax conditioned caption decoder relying on the encoded syntactic and semantic information to produce the syntax customized video caption. Best viewed in color.}
\label{fig:framework}
\end{figure*}

\section{The Proposed Model}

In this section, we propose one novel model to tackle the SCVC task, with the model overview illustrated in Figure~\ref{fig:framework}. Concretely, the proposed model consists of three components, namely the hierarchical sentence syntax encoder, the video semantic encoder and the syntax conditioned caption decoder. We will detail the model and demonstrate our training strategy in the following.

\subsection{Hierarchical Sentence Syntax Encoder}

\label{para:syntax encoder}

To make one video caption imitate the syntactic structure of the given exemplar sentence, how to extract the exemplar sentence syntaxes rather than its semantics is an essential problem. Hence, as shown in the top-right part of Figure~\ref{fig:framework}, we propose a hierarchical sentence syntax encoder as follows.

Specifically, suppose an exemplar sentence $S=[w_1,...,w_n,...,w_N ]$ with $N$ words, in which the $n$-th word consists of $L$ characters $w_n = [w_{n,1},...,w_{n,l},...,w_{n,L}]$. A character-level LSTM, namely ${\rm{LSTM}}^{c}$, is first performed on the sequential characters of each word:
\begin{equation}
\setlength{\abovedisplayskip}{6pt}
\setlength{\belowdisplayskip}{6pt}
    \mathbf{h}^{c}_{n,l}, \mathbf{c}^{c}_{n,l} = {\rm{LSTM}}^{c}(\mathbf{w}^c_{n,l},\mathbf{h}^{c}_{n,l-1}, \mathbf{c}^{c}_{n,l-1}).
\end{equation}
Here $\mathbf{w}^c_{n,l}$ is the embedding of the character $w_{n,l}$ in the word $w_n$, and the corresponding hidden vector $\mathbf{h}^{c}_{n,l}$ will encode the context in the word to characterize the subword features of $w_n$. We average the $L$ character-level hidden vectors, and obtain the syntactic word feature $\mathbf{w}_n =  \rm{Avg}(\mathbf{h}^c_{n,1},...,\mathbf{h}^c_{n,L})$. As words with similar syntactic behaviors, such as part-of-speech (POS), often have similar subword characteristics like suffix or prefix~\cite{hall2014less}, our character-based word feature $\mathbf{w}_n$ will capture the local lexical features to support the sentence level syntax encoding. Subsequently, another word-level LSTM, namely ${\rm{LSTM}}^{w}$, is stacked on ${\rm{LSTM}}^{c}$ to aggregate the syntactic word features:
\begin{equation}
\setlength{\abovedisplayskip}{6pt}
\setlength{\belowdisplayskip}{6pt}
    \mathbf{h}^{w}_{n}, \mathbf{c}^{w}_{n} = {\rm{LSTM}}^{w}(\mathbf{w}_{n},\mathbf{h}^{w}_{n-1}, \mathbf{c}^{w}_{n-1}).
\end{equation}
With such a hierarchical encoding strategy, the syntactic representation $\mathbf{H}^s = [\mathbf{h}^w_1,...,\mathbf{h}^w_{n},...,\mathbf{h}^w_N]$ of the exemplar sentence $S$ is obtained, which will be used to control the sentence syntax of the predicted video caption.

\textbf{Word Replacement Mechanism based on POS Tags.} When conducting experiments, we observe that the sentence syntax encoder often tends to remember the exact words instead of learning the sentence syntactic structure. In order to make the syntax encoder put more emphasis on the sentence syntaxes and lexical word features, we propose a word replacement mechanism based on their POS tags. Specifically, we firstly use the Stanford NLP toolkit~\cite{manning2014stanford} to get the POS tag of each word in the sentence. Then, when feeding the sentence in the syntax encoder, we randomly replace the word in the sentence with the word of the same POS tag, and thus discourage the syntax encoder from memorizing the exact word in the exemplar sentence.

\subsection{Video Semantic Encoder}

For encoding video semantic features, we follow the conventional approaches which first use one pretrained CNN to encode each video frame in the input video $V$, and get a video feature sequence $\mathbf{V}=[\mathbf{v}_1,...,\mathbf{v}_m,...,\mathbf{v}_M]$. Then, an LSTM is leveraged to encode the video contexts, with the aggregated hidden vectors $\mathbf{H}^v = [\mathbf{h}^v_1,...,\mathbf{h}^v_m,...,\mathbf{h}^v_M]$ denoting the video semantic representation, which will further be utilized to provide semantic meanings of the generated video caption.

\subsection{Syntax Conditioned Caption Decoder}

\label{decoder}

In this section, we propose to condition the caption generation process on the encoded exemplar sentence syntaxes to achieve the caption syntax customization. The prediction of each word in the caption is on the basis of the global syntax ``backbone'', and the video semantics are taken as the ``flesh'' to replenish the caption with semantic meanings. Based on this consideration, we establish a two-layer LSTM to decode the caption, with the first layer modeling sentence syntaxes while the second layer fulfilling sentence semantics.

When decoding the $t$-th word in the caption, the first layer decoding LSTM, which we denote as ${\rm LSTM}^{syn}$, takes the concatenation of the embedding of the previous word $\mathbf{e}_{t-1}$ and the attentively summarized syntactic representation $\mathbf{a}^s_{t} = {\rm{Att}}(\mathbf{h}^{syn}_{t-1},\mathbf{H}^s)$ as input, outputting the hidden vector $\mathbf{h}^{syn}_{t}$ to control the sentence syntax at this timestep:
\begin{equation}
\setlength{\abovedisplayskip}{5pt}
\setlength{\belowdisplayskip}{5pt}
\label{LSTM_syntax}
    \mathbf{h}^{syn}_{t}, \mathbf{c}^{syn}_{t} = {\rm{LSTM}}^{syn}([\mathbf{e}_{t-1},\mathbf{a}^s_{t}],\mathbf{h}^{syn}_{t-1}, \mathbf{c}^{syn}_{t-1} \;  \big| \; \mathbf{a}^s_{t}).
\end{equation}
In order to incorporate the syntactic information $\mathbf{a}^s_{t}$ and make it better control the recurrent word decoding procedure, we propose a new Conditional Layer Normalization (${\rm \textbf{CLN}}$) mechanism to modulate $\rm{LSTM}^{syn}$ as shown in the bottom-left part of Figure~\ref{fig:framework}. Specifically, taking $\mathbf{a}_t^s$ as the conditional guidance signal, the computation flow of Eq.~(\ref{LSTM_syntax}) proceeds as:
\begin{equation}
\setlength{\abovedisplayskip}{4pt}
\setlength{\belowdisplayskip}{4pt}
\begin{split}
 [\mathbf{f}_t^{syn},\mathbf{i}_t^{syn},& \mathbf{o}_t^{syn},\mathbf{g}_t^{syn} ] = {\rm \textbf{CLN}} (\mathbf{W}_h^{syn}  \mathbf{h}^{syn}_{t-1} \;  \big| \; \mathbf{a}_t^s) \\ &+ {\rm \textbf{CLN}}(\mathbf{W}_{i}^{syn} [\mathbf{e}_{t-1},\mathbf{a}^s_{t}] \;  \big| \;\mathbf{a}_t^s) + \mathbf{b}^{syn},  \\
\mathbf{c}_t^{syn} &= \sigma(\mathbf{f}_t^{syn}) \odot \mathbf{c}_{t-1}^{syn} + \sigma({\mathbf{i}_t^{syn}}) \odot {\rm tanh}(\mathbf{g}_t^{syn}),  \\
 \mathbf{h}_t^{syn} &= \sigma(\mathbf{o}_t^{syn}) \odot {\rm tanh} \left( {\rm \textbf{CLN}}(\mathbf{c}_t^{syn} \;  \big| \; \mathbf{a}_t^s) \right).
\end{split}
\label{LSTM_CLN}
\end{equation}
Here ${\rm \textbf{CLN}}$ performs the modulation on the LSTM gates $\{\mathbf{f}_t^{syn},\mathbf{i}_t^{syn},\mathbf{o}_t^{syn},\mathbf{g}_t^{syn}\}$ and cell $\mathbf{c}_t^{syn}$, which leverages $\mathbf{a}_t^s$ to conditionally scale and shift the layer normalized vectors:
\begin{equation}
\setlength{\abovedisplayskip}{3pt}
\setlength{\belowdisplayskip}{3pt}
{\rm \textbf{CLN}}(\mathbf{x}  \;  \big| \; \mathbf{a}^s_{t}) = f_{\gamma}(\mathbf{a}_t^s) \cdot \frac{\mathbf{x}-\bm{\mu}(\mathbf{x})}{\bm{\sigma}(\mathbf{x})} + f_{\beta}(\mathbf{a}_t^s).
\label{CLN}
\end{equation}

The scaling and shifting vectors $f_{\gamma}(\mathbf{a}_t^s)$ and $f_{\beta}(\mathbf{a}_t^s)$ are generated by two independent multi-layer perceptrons (MLPs) with $\mathbf{a}_t^s$ as the input, and the three ${\rm \textbf{CLN}}$s in Eq.~(\ref{LSTM_CLN}) are independent with each other and do not share weights. By equipping ${\rm \textbf{CLN}}$s, the syntactic representation $\mathbf{a}_t^s$  is expected to affect the update procedure of ${\rm LSTM}^{syn}$, thereby achieving the caption syntax customization.

The second layer LSTM in our syntax conditioned caption decoder, which we denote as ${\rm LSTM}^{sem}$, has the same architecture as ${\rm LSTM}^{syn}$. Specifically, ${\rm LSTM}^{sem}$ takes the concatenation of the syntax hidden vector $\mathbf{h}^{syn}_{t}$ and the attentively summarized video semantic representation $\mathbf{a}^v_{t} = {\rm{Att}}(\mathbf{h}^{sem}_{t-1},\mathbf{H}^v)$ as inputs, and is conditionally controlled under the video semantics:
\begin{equation}
\setlength{\abovedisplayskip}{5pt}
\setlength{\belowdisplayskip}{5pt}
    \mathbf{h}^{sem}_{t}, \mathbf{c}^{sem}_{t} = {\rm{LSTM}}^{sem}([\mathbf{h}_{t}^{syn},\mathbf{a}^v_{t}],\mathbf{h}^{sem}_{t-1}, \mathbf{c}^{sem}_{t-1} \;  \big| \; \mathbf{a}^v_{t}).
\end{equation}
Here, $\mathbf{a}^v_{t}$ is taken as the condition guidance signal to intensify the video contents in the word decoding procedure, making the predicted caption better preserve the video semantics. Through the above two-layer LSTM modeling, the words in the caption will be sequentially predicted based on the hidden vectors $\mathbf{H}^{sem}=[\mathbf{h}^{sem}_{1},..,\mathbf{h}^{sem}_{t},...,\mathbf{h}^{sem}_{T}]$.

\subsection{Training Strategy}
\label{training_para}

As we do not have the syntax customized groundtruth video captions, we propose to leverage the pairwise video captioning data in available datasets as well as our collected auxiliary exemplar sentence corpus to train our proposed model.

\subsubsection{Training with Pairwise Video Captioning Data}

\begin{figure}
\includegraphics[width=1.0\columnwidth]{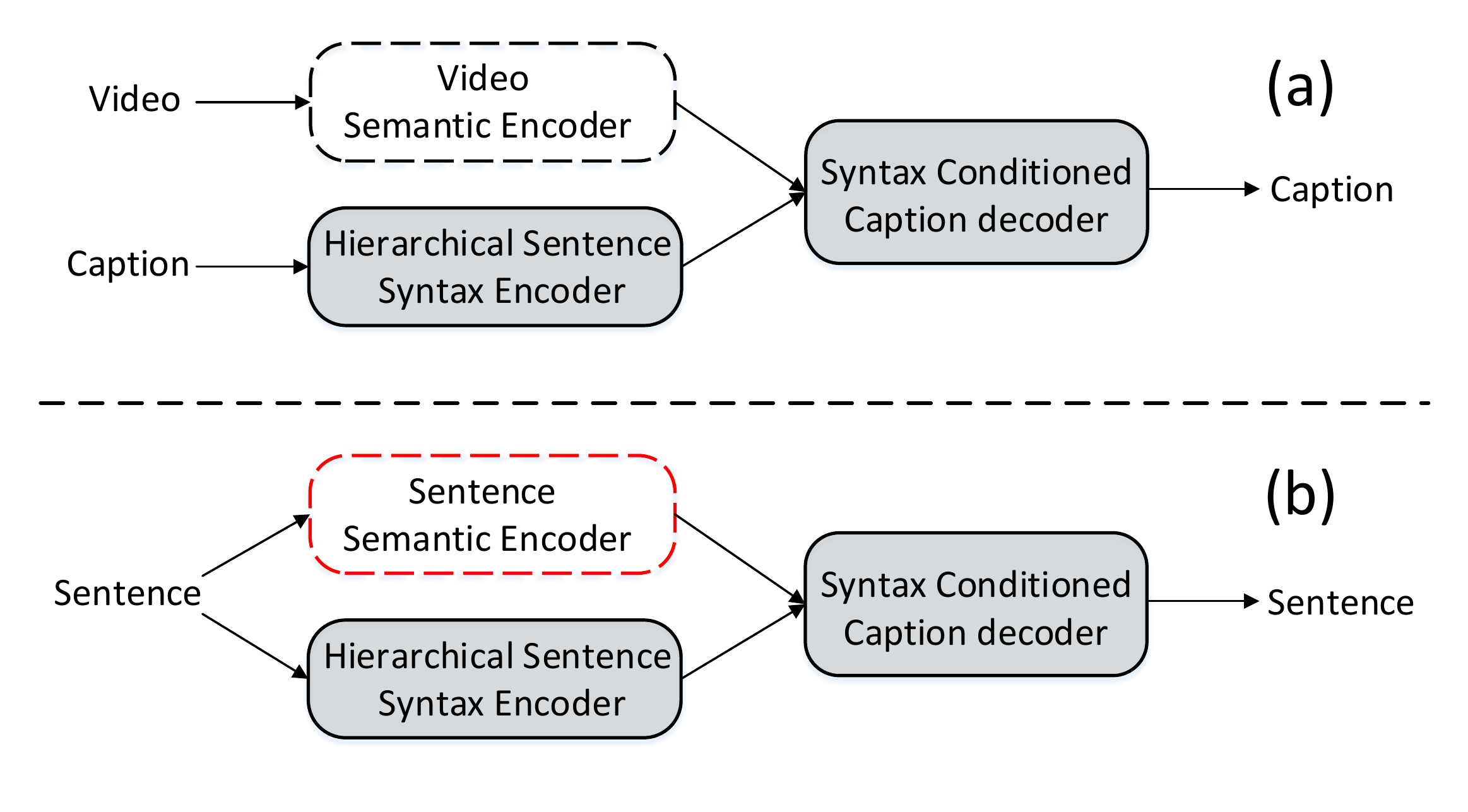} 
\caption{ Our proposed training strategy leverages (a) the pairwise video captioning data and (b) our collected auxiliary exemplar sentences to train our proposed model. The shadowed blocks are shared when training.}
\label{fig:training}
\end{figure}

As illustrated in Figure~\ref{fig:training}(a), the pairwise video captioning data is utilized for training the proposed model, with the groundtruth caption also being taken as the exemplar sentence. As such, the sentence syntax of the caption is extracted and then coupled with the video semantics to predict the caption. For the training of this syntax customized video captioning procedure,  two loss terms are introduced as the learning objective, \textit{i.e.}, the syntactic loss and the semantic loss.

\textbf{Syntactic Loss.} In our syntax customized video captioning, the syntactic structure of the generated caption should resemble that of the exemplar sentence. We use the Stanford NLP toolkit~\cite{Manning2014The} to extract the constituency parse tree of the exemplar sentence (here is the input video caption), which is presented as a bracket syntactic token sequence. Then we collect the hidden state vectors of the $\rm{LSTM}^{syn}$ as $\mathbf{H}^{syn}$=$[\mathbf{h}^{syn}_1,...,\mathbf{h}^{syn}_t,...,\mathbf{h}^{syn}_T]$, feed them into a basic decoding model such as~\cite{yao2015describing}, and predict the corresponding syntactic token sequence of the input exemplar sentence. Accordingly, the syntactic loss can be defined as:
\begin{equation}
\setlength{\abovedisplayskip}{6pt}
\setlength{\belowdisplayskip}{5pt}
    L_{v,c}^{syn} = - {\rm log} P(C^{syn}|\mathbf{H}^{syn};V,C).
\end{equation}
$L_{v,c}^{syn}$ is realized by the typical negative log-likelihood loss. $V$ denotes the input video, and $C$ denotes its accompanying caption in the dataset, which is also used as the exemplar sentence. $C^{syn}$ indicates the extracted syntactic token sequence of the input caption $C$.

\textbf{Semantic Loss.} Since the generated caption should express the semantic meaning of the video, we collect the hidden vectors $\mathbf{H}^{sem} = [\mathbf{h}^{sem}_1,...,\mathbf{h}^{sem}_t,...,\mathbf{h}^{sem}_T]$ of the $\rm{ LSTM}^{sem}$ that directly serves for caption decoding, and adopt the negative log-likelihood loss for semantic supervision with $C$:
\begin{equation}
\setlength{\abovedisplayskip}{6pt}
\setlength{\belowdisplayskip}{5pt}
    L_{v,c}^{sem} = - {\rm log} P(C|\mathbf{H}^{sem};V,C).
\end{equation}

\subsubsection{Training with An Exemplar Sentence Corpus}
\label{sec:sentenceloss}

Since the syntactic structures of the captions in existing video captioning datasets are simple and monotonous, only relying on them to train our model will limit its ability on expression variability. To this end, we collect a large number of syntax-varied exemplar sentences to enhance our model's learning ability. Meanwhile, as shown in Figure~\ref{fig:training}(b), we propose a sentence semantic encoder to replace the video semantic encoder in our model. Doing so will make one single sentence can also be autoencoded with the yielded architecture, and further help the training of the shared hierarchical sentence syntax encoder and syntax conditioned caption decoder.

Concretely, the sentence semantic encoder relies on one fully-connected layer to encode the glove embedding~\cite{pennington2014glove} of each word, and the obtained word feature sequence will be taken as the sentence semantic representation. By feeding the semantic and syntactic representations of the sentence to the syntax conditioned caption decoder, we follow the same procedure in Sec.~\ref{decoder} to reconstruct the input sentence. The corresponding syntactic and semantic losses can also be defined:
\begin{equation}
\setlength{\abovedisplayskip}{6pt}
\setlength{\belowdisplayskip}{5pt}
\begin{split}
    L_{s,s}^{syn} &= - {\rm log} P(S^{syn}|\mathbf{H}^{syn};S,S), \\
    L_{s,s}^{sem} &= - {\rm log} P(S|\mathbf{H}^{sem};S,S).
\end{split}
\end{equation}
Here $S$ denotes the sentence we feed to the sentence autoencoding procedure, which can be our collected exemplar sentence $E$ or the caption $C$ in the existing video captioning datasets. $S^{syn}$ denotes the parse tree token sequence of $S$.

\subsubsection{Overall Training Objective}

The above training objective considers either feeding a (video, caption) pair or one single sentence to our model. However, there is also one case of taking one exemplar sentence $E$ and one video $V$ as the inputs to our model. Although we do not have groundtruth caption to support the $L^{sem}$ loss in this case, the $L^{syn}$ loss can still be involved in training since the parse tree sequence $E^{syn}$ of the exemplar sentence can be pre-extracted. Hence, we can obtain an additional syntactic loss as follows:

\begin{equation}
L^{syn}_{v,e} =  - {\rm log} P(E^{syn}|\mathbf{H}^{syn};V,E).
\end{equation}
Summing up the above terms, the overall training objective of our model is defined as:
\begin{equation}
L = L_{v,c}^{syn}+L_{v,c}^{sem}+L_{s,s}^{syn}+L_{s,s}^{sem}+L^{syn}_{v,e}.
\end{equation}

\section{Experiments}

\newcommand{\tabincell}[2]{\begin{tabular}{@{}#1@{}}#2\end{tabular}}
\begin{table*}[!t]
  \centering
  \begin{threeparttable}
  \caption{Performance comparisons on MSRVTT and ActivityNet Captions.}
  \label{tab:performance_comparison}
    \begin{tabular}{ccccccccc}
    \toprule
    \multirow{3}{*}{Method}
    &\multicolumn{3}{c}{MSRVTT}
    &\multicolumn{3}{c}{ActivityNet Captions}
    \cr
    \cmidrule(lr){2-4} \cmidrule(lr){5-7}
    &\tabincell{c}{TED$\downarrow$} 
    &\tabincell{c}{COS$\uparrow$} 
    &\tabincell{c}{perplexity$\downarrow$}     
    &\tabincell{c}{TED$\downarrow$} 
    &\tabincell{c}{COS$\uparrow$} 
    &\tabincell{c}{perplexity$\downarrow$} 
    \cr
    \midrule
    ExemplarOnly
    &0.00 &0.5238 &4.92 
    &0.00 &0.5784 &4.91  \cr

    Seq2Seq
     &15.91 &0.6949 &3.51
     &20.12 &0.7317 &2.58  \cr

     Template
    &4.88 &0.6595 &7.75
    &2.97 &0.6945 &7.58  \cr

    GFN
    &16.08 &0.6963 &3.86 
    &20.86 &0.7467 &2.80  \cr

    Ours
    &5.44 &0.6892 &5.64 
    &5.52 &0.7113 &6.61  \cr
    \bottomrule
    \end{tabular}
    \end{threeparttable}
\end{table*}

\subsection{Dataset and Exemplar Sentence Collection}

We conduct our syntax customized video captioning experiments on the MSRVTT \cite{xu2016msr} and ActivityNet Captions \cite{krishna2017dense} datasets. We pair each video in these two datasets with 20 different syntax-varied exemplar sentences that are originally collected in the Shutterstock Image Description Corpus \cite{Feng2018Unsupervised}. Accordingly, 20 different syntax customized video captions will be generated by our proposed model. The details of the datasets and our exemplar sentence collection are as follows.

\noindent \textbf{MSRVTT \cite{xu2016msr}}. The MSRVTT is a large-scale dataset for video captioning. It contains 10k video clips and each video clip is accompanied with 20 human-edited English sentence descriptions, resulting in 200K video-caption pairs in total. Following the existing works, we use the public data split in our experiments, i.e., 6513 videos for training, 497 for validation, and 2990 for testing.

\noindent \textbf{ActivityNet Captions \cite{krishna2017dense}}. The ActivityNet Captions is a benchmark dataset proposed for dense video captioning. There are 20K untrimmed videos in total, and each video has several annotated segments with starting and ending times as well as the associated captions. Overall it contains 10,024 videos for training, 4,926 videos for validation and 5,044 for testing. Since we do not perform dense video captioning in this work, we split the caption-paired segments from the training and validation videos, and perform video captioning on them. In this way, 54,926 segment-caption pairs are collected, where 37,421 segments from the public training set are used for training, and 17,505 segments from the validation set are used for testing.

\noindent \textbf{Exemplar Sentence Collection.} The exemplar sentences in our work should meet the following two requirements. (1) The sentences should be human-edited. (2) The sentences should have various syntactic structures with no other restrictions. We find that the recently collected Shutterstock Image Description Corpus  \cite{Feng2018Unsupervised} is quite appropriate. Specifically, the Shutterstock Image Description Corpus is collected by crawling the image descriptions from Shutterstock for the unsupervised image captioning research. Shutterstock is an online stock photography website, which provides hundreds of millions of royalty-free stock images. All the image descriptions are written by image composers, and have diverse sentence syntactic forms.  We download the collected 2,322,628 image descriptions in the Shutterstock Image Description Corpus, and filter the descriptions that are less than 8 words or longer than 30 words, obtaining a total of 761,582 exemplar sentences. For each video/segment in the MSRVTT and ActivityNet Captions, we randomly choose 20 descriptions as its exemplar sentences for our syntax customized video captioning task.

\subsection{Evaluation Metrics}
To comprehensively evaluate the predicted syntax customized video captions, we refer to the research in the relating text style transfer task~\cite{fu2018style,jin2020deep} and conduct the objective evaluation from  the \textbf{syntactic}, \textbf{semantic} and \textbf{fluency} aspects as follows.

\noindent \textbf{Syntactic Evaluation.} To evaluate whether our generated captions comply with the syntactic structures of the exemplar sentences, we directly compare their constituency parse tree by computing the syntactic Tree Edit Distance (TED)  \cite{zhang1989simple} between them after removing word tokens. Smaller TED value means higher syntactic similarity.

\noindent \textbf{Semantic Evaluation.} For each sentence, we first remove the stop words and then take the average glove word embeddings of the remaining words as its sentence semantic features. In the semantic feature space, the average cosine similarity (COS) between the syntax customized video caption and all the originally groundtruth video captions in the dataset is used to evaluate their semantic coherence.

\noindent \textbf{Sentence Fluency. } Besides syntactic and semantic evaluation, another important aspect is the sentence fluency of the generated syntax customized video captions. We use a pre-trained language model to measure the perplexity of generated captions as the fluency score. A state-of-the-art BERT model \cite{Wolf2019HuggingFacesTS} trained on a large scale lower case English dataset is used for the evaluation.

\subsection{Implementation Details}
To represent videos, we leverage the Inception-Resnet-v2 network \cite{Szegedy2016Inception} pretrained on the ILSVRC-2012-CLS image classification dataset \cite{Russakovsky2015ImageNet} to extract a 1,536 dimensional feature vector for each frame. Videos in the MSRVTT and ActivityNet Captions datasets are represented with evenly spaced 30
and 100 features, respectively. Shorter videos of less than 30 or 100 features are padded with zero vectors. The word embedding size and all the LSTMs’ hidden sizes are set as 256. Adam \cite{KingmaAdam} optimizer is used for training. For the word replacement mechanism, the probability to replace the word with their same POS-tagged word is 0.7. The model is implemented with PyTorch \cite{paszke2019pytorch}.

For getting the syntactic token sequence, we first extract the constituency parse tree of the input exemplar sentence with Stanford NLP parser \cite{Manning2014The}. For example, the parse tree of the sentence  ``\texttt{ \small a short clip of news on a white background}'' is ``\texttt{ \small (ROOT (NP (NP (NP (DT) (JJ) (NN)) (PP (IN) (NP (NN)))) (PP (IN) (NP (DT) (JJ) (NN)))))}''. We regard the sentence parse as a syntactic token sequence, where each element such as ``\texttt{\small ROOT}'', ``\texttt{\small NN}'' or ``\texttt{\small (}'' is an independent token in the sequence.  Such a syntactic token sequence will be used for the syntactic loss as stated in Sec.~\ref{training_para}.

\subsection{Compared Methods}

Since the proposed syntax customized video captioning is a new task, there is no baseline method for direct comparisons. In this section, we compare our proposed model with the following methods:

\begin{itemize}
    \item The \textbf{ExemplarOnly} method which directly outputs the given exemplar sentence as the video caption while does not consider the video contents. 
    \item The conventional sequence-to-sequence (\textbf{Seq2Seq}) video captioning method \cite{yao2015describing}, which only predicts one single caption for one video while does not consider the sentence syntax.
    \item The \textbf{Template} method, which firstly detects visual concepts~\cite{Fang_2015_CVPR} from each video to generate a bag of video-related words. Then every content word in the exemplar sentence is simply replaced by one video-related word from the bag that has the same part-of-speech as the removed word.
    \item The Gated Fusion Network (\textbf{GFN}) for video captioning with POS sequence guidance \cite{wang2019controllable}. We replace their intermediately sampled POS sequences from videos with the POS sequences of the given exemplar sentences, and attain the corresponding syntax customized video captions.
\end{itemize}

\subsection {Performance Comparison}

\subsubsection{Quantitative Evaluation}

Table \ref{tab:performance_comparison} compares our proposed model with the baseline methods in terms of the objective evaluation metrics introduced above.
The ExemplarOnly method directly outputs exemplar sentences without considering video contents, making the exemplar and output sentence structures exactly the same and TED scores as 0.0, whereas the semantic coherence to the video is much lower.
The Seq2Seq model mainly focuses on describing video contents while neglecting the given exemplar sentence syntaxes, so it gets higher semantic COS scores but inferior TED scores.
The Template method simply fills the detected content words from videos in the exemplar sentences while does not consider the global sentence meanings. Doing so makes the generated captions fit the original exemplar syntaxes very well but the captions are not fluent and get high (poor) perplexity scores. Meanwhile, the semantic scores are also lower.
Although GFN attempts to leverage POS features to guide video captioning, its controllability of caption syntax is weak and results in high TEDs between exemplar sentences and captions. The main focus of GFN is still to improve the semantic accuracy of video captions and keep their fidelity with the training corpus.
Since the Seq2Seq and GFN models often generate simple sentence descriptions with common sentence forms, their perplexity scores are correspondingly lower. The collected exemplar sentences are human-edited, which often take more various and complex syntaxes and are free in grammar. Thus, the perplexity scores of the ExemplarOnly method are relatively higher. Based on these complicated exemplars, the yielded Template and Ours method will also inevitably present higher perplexity scores.

Our proposed model achieves much better TED scores and comparable COS scores than the Seq2Seq and GFN models, which shows that our method can comply with the exemplar sentence syntaxes while describing the video semantics. Also, imitating the complex exemplar sentences does not significantly affect the sentence fluency of our generated captions, with the perplexity scores increasing in a reasonable range compared to the ExemplarOnly method. 

\begin{table}
  \centering
  \caption{ Human evaluation of captioning.}
    \begin{tabular}{cccc}
    \toprule
    Method
    &\tabincell{c}{Syntactic\\Score} &\tabincell{c}{Semantic\\Score} &\tabincell{c}{Fluency\\Score}  \cr

    \midrule
    ExemplarOnly
    &1.0000 &0.0040 &0.9260 \cr

    Seq2Seq 
    &0.0043 &0.5945 &0.9938 \cr

    Template
    &0.9205 &0.5322 &0.6530  \cr

    GFN 
    &0.0058 &0.6153 &0.9753 \cr

    Ours
    &0.8895 &0.5773 &0.8560 \cr
    \bottomrule
    \end{tabular}
\end{table}

\begin{table}
    \caption{ Captioning diversity evaluation.}
    \centering
    \resizebox{0.48\textwidth}{!}{
    \begin{tabular}{ccccc}
    \toprule
    \multirow{2}{*}{Method}&
    \multicolumn{2}{c}{MSRVTT}&
    \multicolumn{2}{c}{ActivityNet Captions}\cr
    \cmidrule(lr){2-3} \cmidrule(lr){4-5}
    &\tabincell{c}{LSA}
    &\tabincell{c}{Self-CIDER}
    &\tabincell{c}{LSA}
    &\tabincell{c}{Self-CIDER}
    \cr
    \midrule
    ExemplarOnly
    &0.7437 &0.9737
    &0.7431 &0.9741   \cr

    Seq2Seq \cite{yao2015describing}
    &0.0000 &0.0000
     &0.0000 &0.0000  \cr

    Template
    &0.5073 &0.7940
    &0.5067 &0.7943   \cr

    GFN \cite{wang2019controllable}
    &0.2194 &0.3225
    &0.2553 &0.4241 \cr

    Ours
    &0.6150 &0.8814
    &0.6973 &0.9365 \cr

    \bottomrule
\end{tabular}
}
\end{table}

\begin{table*}[!t]
  \centering
  \begin{threeparttable}
  \caption{Evaluation results in the conventional captioning metrics (\%).}
  \label{tab:performance_conventional_metric}
    \begin{tabular}{ccccccccc}
    \toprule
    \multirow{2}{*}{Method}&
    \multicolumn{4}{c}{MSRVTT}&\multicolumn{4}{c}{ActivityNet Captions}\cr
    \cmidrule(lr){2-5} \cmidrule(lr){6-9}
    &\tabincell{c}{B@4} 
    &\tabincell{c}{CIDEr}
    &\tabincell{c}{ROUGE$^{*}$}
    &\tabincell{c}{METEOR}
    &\tabincell{c}{B@4} 
    &\tabincell{c}{CIDEr}
    &\tabincell{c}{ROUGE$^{*}$}
    &\tabincell{c}{METEOR}
    \cr
    \midrule

    Seq2Seq \cite{yao2015describing}
    &37.20 &40.24 &58.64 &26.39
    &3.36 &23.92 &20.44 &8.93  \cr

    Template
    &0.86 &4.87 &23.81 &12.59
    &0.11 &6.26 &10.15 &4.71   \cr

    GFN \cite{wang2019controllable}
    &38.74 &43.90 &59.42 &27.11
    &4.65 &32.87 &21.65 &10.24   \cr

    Ours
    &3.29 &11.45 &28.45 &15.33 
    &0.20 &8.18 &9.73 &4.50  \cr
    \bottomrule
    \end{tabular}
    \end{threeparttable}
  \begin{tablenotes}
  \footnotesize
    \item *: Here ROUGE indicates ROUGE\_L.
     \end{tablenotes}
\end{table*}

\begin{figure*}[!t]
\centering
\includegraphics[width=1.0\textwidth]{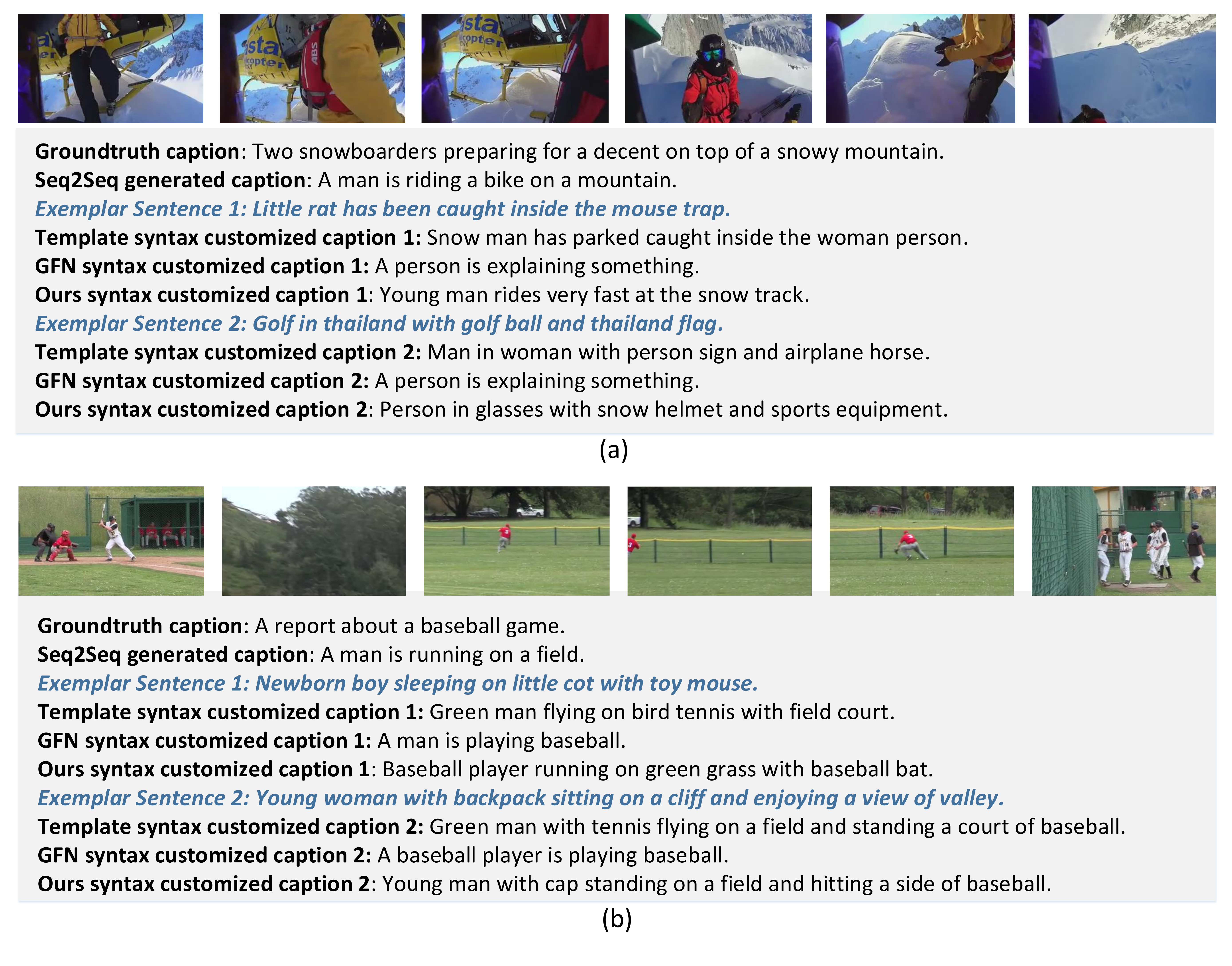} 
\caption{ Qualitative results for the syntax customized video captioning. For each case, we provide the video, the  groundtruth caption, and the caption generated by the Seq2Seq model. Two exemplar sentences are given to each video. The Template, GFN, and our predicted syntax customized video captions are also present correspondingly.}
\label{fig:qualitative_results}
\end{figure*}

\subsubsection{Qualitative Evaluation} 

Figure \ref{fig:qualitative_results} shows the qualitative results for the syntax customized video captioning task. The Seq2Seq model can generate only one simple caption for one video, which is monotonous and bears high resemblance to the training corpus. The Template method which simply replaces main content words in the exemplar sentences make the generated captions lack of fluency. The GFN model cannot capture sentence syntaxes and produces similar or the same captions for different exemplar sentences. Our generated captions comply with the exemplar sentence syntaxes well, and more concrete video contents like  ``snow helmet'' and ``green grass'' can also be described in the captions more expressively. Meanwhile, the two different syntax customized video captions further provide us the intuition of diverse video captioning. 

\begin{figure*}
\setlength{\abovecaptionskip}{0.cm}
\setlength{\belowcaptionskip}{-0.1cm}
    \centering
    \includegraphics[width=0.9\textwidth]{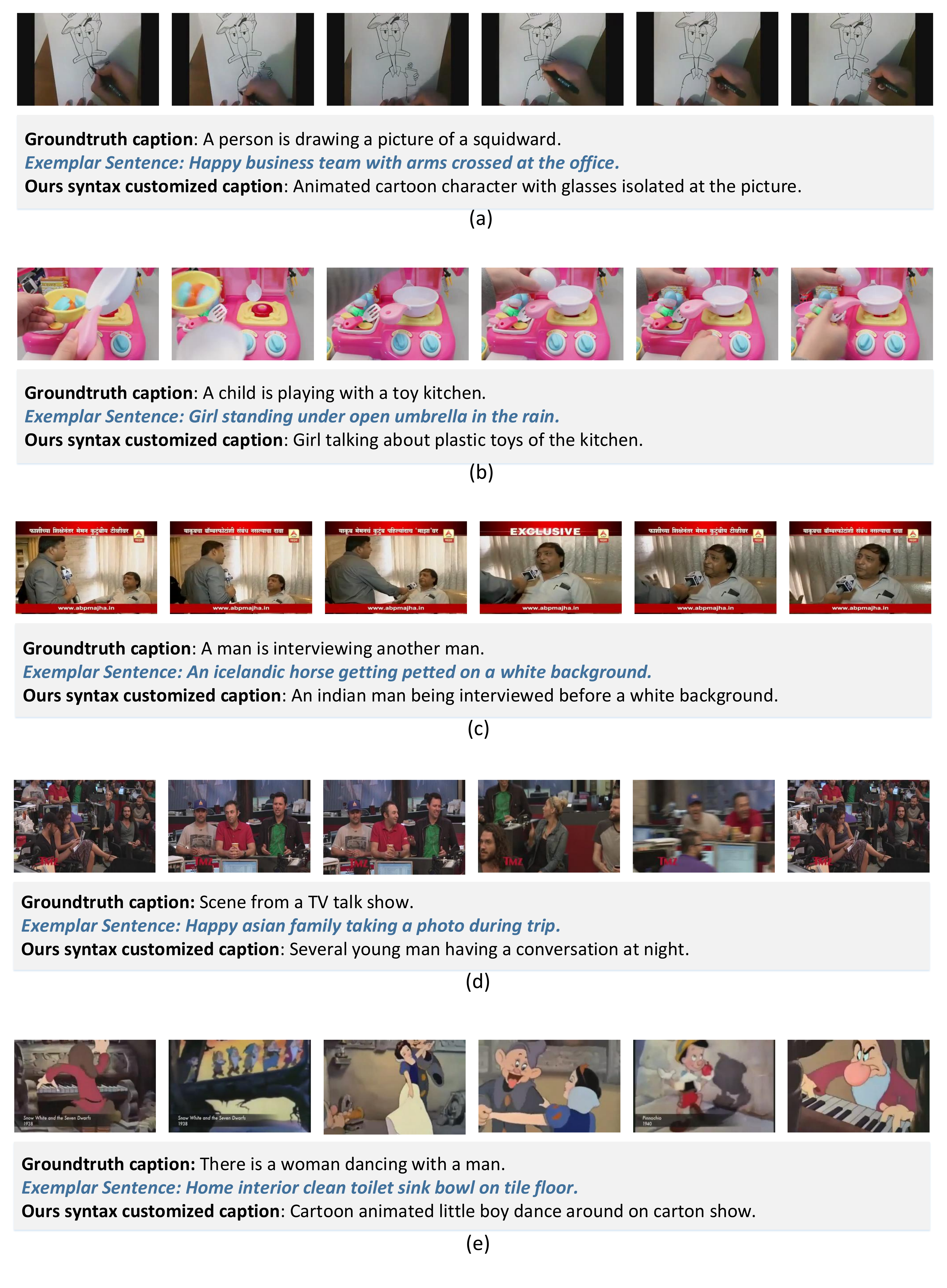} 
    \caption{More qualitative results on the MSRVTT dataset. For each video, we provide one groundtruth caption, one exemplar sentence, and the syntax customized video caption predicted by our proposed model. }
    \label{fig:msrvtt_supp_qualitative_1}
\end{figure*}

\begin{figure*}
\setlength{\abovecaptionskip}{0.cm}
\setlength{\belowcaptionskip}{-0.1cm}
    \centering
    \includegraphics[width=0.9\textwidth]{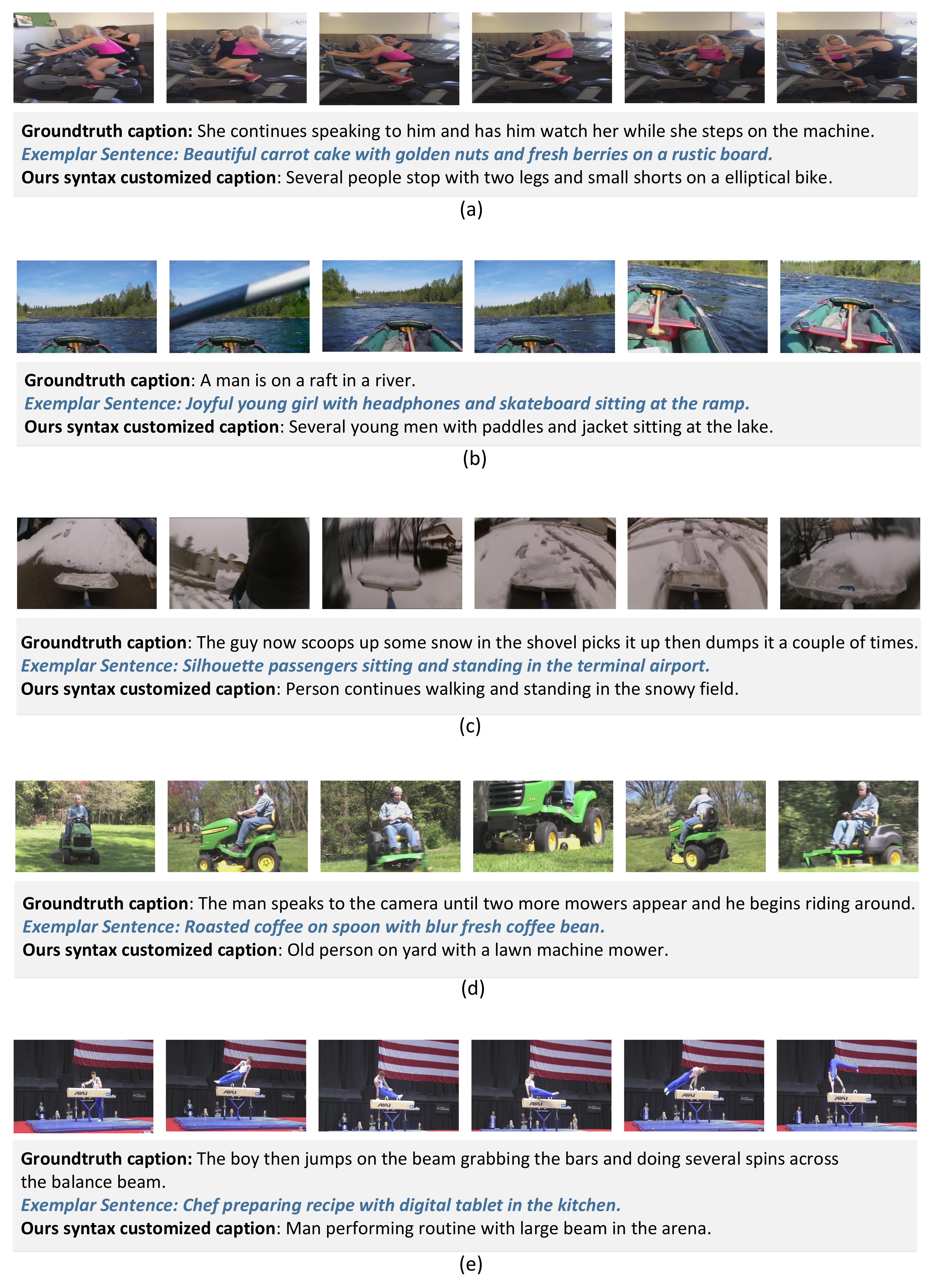} 
    \caption{More qualitative results on the ActivityNet Captions dataset. For each video, we provide one groundtruth caption, one exemplar sentence, and the syntax customized video caption predicted by our proposed model. }
    \label{fig:anet_supp_qualitative_1}
\end{figure*}

To better demonstrate the effectiveness of our proposed method on syntax customized video captioning, we further provide more qualitative results of Ours method in Figure~\ref{fig:msrvtt_supp_qualitative_1} and Figure~\ref{fig:anet_supp_qualitative_1} at the end of the paper. It can be observed that on different types and contents of videos, our generated captions can not only present the video semantics accurately, but also  well imitate the syntaxes of the given exemplar sentences.

\subsubsection{Human Evaluation} 

Besides the quantitative and qualitative evaluations, we also conduct a human assessment of the generated captions. The human evaluators were asked to rate the three aspects of the generated captions --- syntax similarity with the exemplar sentence, semantic coherence with the video, and sentence fluency. Each rating aspect is graded in three scales \{0.0, 0.5, 1.0\}, and the higher, the better. We randomly chose 200 captions generated by each compared method from the test set and invited 10 evaluators (5 males and 5 females) to grade them.  The average rating scores are presented in Table 2. These scores are in agreement with the above quantitative evaluation. Although the complex exemplar sentence syntaxes may influence the sentence fluency of our generated captions, they are still realistic and can describe the video semantic contents meanwhile imitating the exemplar sentence syntaxes.

\subsubsection{Captioning Diversity Evaluation}
Since we collect 20 different exemplar sentences for each video, we will accordingly get 20 different syntax customized captions with the proposed model. To measure the diversity among these generated captions, we follow the evaluation metrics introduced in \cite{wang2019describing}, and report the LSA and Self-CIDEr based diversity scores in Table 3. The collected exemplar sentences are independent of each other, and it is evident that the diversity scores of the ExemplarOnly method are fairly high and can be seen as upper bounds in this diversity evaluation experiment. The conventional Seq2Seq model can only generate one single caption for one video, and thus it cannot achieve diverse video captioning and gets all the scores as 0.0. The GFN method can vary the video captions to some extent, while the captioning diversity remains limited. The Template method and Ours model can generate diverse captions to describe the video contents, and Ours model even gets comparable LSA and Self-CIDER scores to other state-of-the-art methods reported in \cite{wang2019describing}. Hence, the results verify that our proposed model can indeed strengthen the diversity of video captioning.

\begin{figure*}[!t]
\centering
\includegraphics[width=0.9\textwidth]{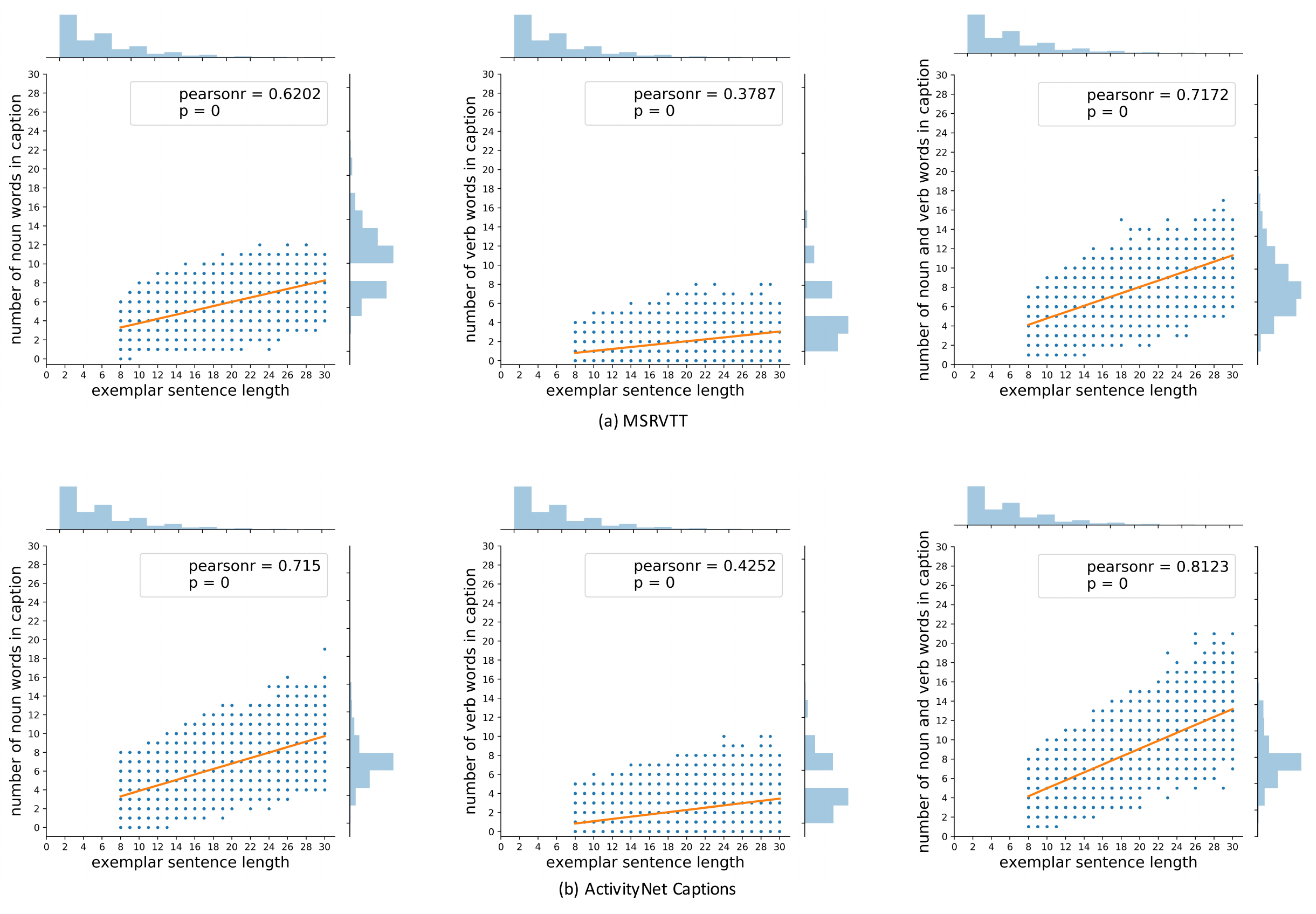}
\caption{The illustration of correlation between the number of object (verb, noun) words in captions and the exemplar sentence lengths. In each scatter map, the x-axis indicates the lengths of exemplar sentences, and the y-axes indicate the number of noun, verb, and noun+verb words in our generated syntax customized video captions, respectively.}
\label{fig:object_num}
\end{figure*}

\subsection{The Influence of Exemplar Sentence Length}

In our experiments, we find that given different exemplar sentences, there can be different object descriptions coming from the video. Therefore, it is also interesting to study the effect of exemplar sentence length to the number of different object words in the generated captions. Therefore, we first count the different verb and noun words in each generated syntax customized video caption of our model. Then, we plot 6 scatter maps in Fig.~\ref{fig:object_num}, where the top three scatters are for MSRVTT dataset and the bottom three are for ActivityNet Captions dataset. The x coordinate of each point in the scatter map means the length of the given exemplar sentence, and the y coordinate means the number of different noun, verb, and noun+verb words (we call them objects in general) in the corresponding syntax customized video caption, respectively. The pearson correlation coefficient between the exemplar sentence length and the object number is also presented in the top-right part of each subfigure. It can be observed that the object number is positively correlated to the exemplar sentence length, no matter for verbs, nouns or their union. Such results indicate that when we provide longer exemplar sentences, more objects will be incorporated in the generated syntax customized captions, making more representative and diverse video descriptions.

\subsection{Discussions on the Conventional Metrics}

The generated syntax customized video captions one the one hand should semantically describe the video contents, on the other hand should syntactically follow the given exemplar sentences, thus making the caption syntactic structures be greatly changed in the syntax imitation procedure. As such, our generated syntax customized video captions will be greatly different from the original groundtruth captions (in the MSRVTT and ActivityNet Captions datasets) on n-gram characteristics and detail wording. Since conventional captioning evaluation metrics, such as BLEU, take the n-gram similarity between the predicted and groundtruth captions as a basic measurement, they are not very appropriate for evaluating our syntax customized video captioning task. 

In this section, we still provide the evaluation results in the conventional video captioning metrics in Table~\ref{tab:performance_conventional_metric} for reference. As expected, our proposed model does not achieve high performances in these metrics, while the Seq2Seq and GFN models get good results because they are still focusing on fitting the detail wording patterns in the training corpus. The different evaluation results between the conventional captioning metrics and our human ratings in the main paper also indicate the limitation of the conventional metrics in evaluating syntax-varied captions.

\begin{figure*}[!t]
\centering
\includegraphics[width=1.0\textwidth]{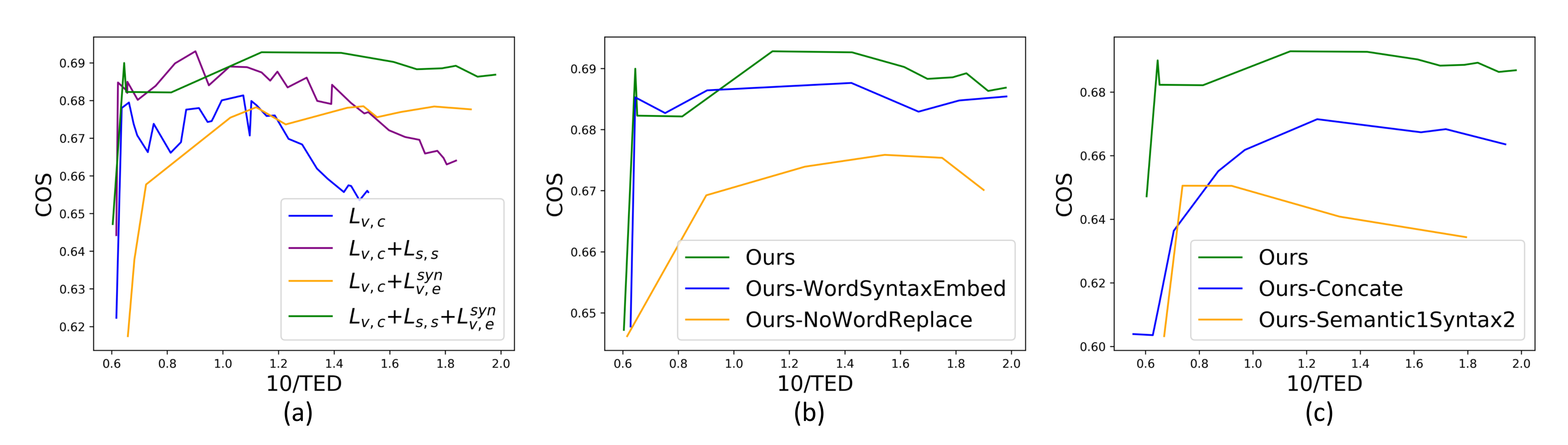} 
\caption{In this figure, we present the ablation studies of the proposed model by examining the contributions on (a) the loss components, (b) the sentence syntax encoding, and (c) the caption decoding. }
\label{all_ablation_studies}
\end{figure*}

\subsection{Ablation Studies}

\subsubsection{Model Contribution Examination}

In this section, we perform three groups of ablation studies on the MSRVTT dataset to examine the contributions of our proposed model, with the results shown in Figure \ref{all_ablation_studies}. Each point in this figure represents the performance of the model on the validation set during the training process. The x-axis indicates the 10/TED scores, and the y-axis denotes the COS scores. Since lower TED and higher COS means better performance, models at the top right are more desirable.

\textbf{Loss Components.} Comparing the model with only the $L_{v,c}$=$L_{v,c}^{syn}$+$L_{v,c}^{sem}$ loss and the model additionally considering the $L_{s,s}$=$L_{s,s}^{syn}$+$L_{s,s}^{sem}$ loss in Figure~\ref{all_ablation_studies}(a), we can observe that training with an exemplar sentence corpus as introduced in Sec.~\ref{sec:sentenceloss} can augment the training data and thereby improve the model performance. By considering both $L_{v,c}$ and $L_{v,e}^{syn}$ terms, the performance curve rises steadily without fierce fluctuation. It shows that even without paired groundtruth syntax customized captions, merely providing supervision on sentence syntaxes can also help the model imitate the syntactic structures of the given exemplar sentences. By combining all the above terms, our proposed model achieves the best performances.

\textbf{Sentence Syntax Encoding.} Ours-WordSyntaxEmbed model in Figure \ref{all_ablation_studies}(b) drops the character-level LSTM, and only keeps the word-level LSTM to encode the sentence syntactic structure. Ours-NoWordReplace model does not adopt the word replacement mechanism as we stated in Sec.~\ref{para:syntax encoder}. The performance superiority of the full model (Ours) over these two ablation models verifies the benefits of the character-level LSTM in encoding subword features and the word replacement mechanism. These designs can help our model capture exemplar sentence syntaxes effectively and filter out unnecessary sentence semantics, which is very crucial for syntax customized video captioning.

\textbf{Caption Decoding.} Ours-Concate model in Figure~\ref{all_ablation_studies}(c) simply concatenates sentence syntactic and video semantic representations and takes them as the inputs to the decoding LSTM without the proposed CLN processing. We can observe that the full model (Ours) achieves higher COS scores than Ours-Concate, which shows that equipping CLN in the decoding LSTM is beneficial to video semantic preservation when performing syntax customization. In Ours-Semantic1Syntax2 model, we feed video semantic representation to the first layer LSTM and sentence syntactic representation to the second. Such a semantic-first syntax-follow architecture is significantly inferior to our proposed model. The reason mainly dues to that using syntactic information to guide the word prediction in the second layer LSTM will hinder it from choosing appropriate words coherent to the video semantics.

\subsubsection{The Influence of Word Replacement Probability}

\begin{table}[!htbp]
  \centering
  \caption{Model performance with different word replacement probability.}
    \begin{tabular}{cccc}
    \toprule
    Method
    & TED$\downarrow$ 
    & COS$\uparrow$ 
    & perplexity$\downarrow$ 
    \cr

    \midrule
    Ours-0.1
    &4.28 &0.6535 &6.82  \cr

    Ours-0.3
    &4.81 &0.6643 &6.65  \cr

    Ours-0.5
    &4.72 &0.6724 &6.44  \cr

    Ours-0.7
    &5.44 &0.6892 &5.64  \cr

    Ours-0.9
    &6.14 &0.6878 &5.61 \cr
    \bottomrule
    \end{tabular}
    \label{tab:probability}
\end{table}

For the word replacement mechanism proposed in Sec.~\ref{para:syntax encoder}, we have mentioned that we will randomly replace the word in the exemplar sentence with the word of the same POS tag with specific probability. In this section, we further  investigate the influence of the word replacement probability to the model performance, by setting the probability value to \{0.1, 0.3, 0.5, 0.7, 0.9\} and evaluating the corresponding models in the MSRVTT dataset. The model performances are shown in Table~\ref{tab:probability}.

It can be observed from Table~\ref{tab:probability} that if the replacement probability is small (\textit{e.g.}, 0.1, 0.3, 0.5), the models get lower COS semantic scores and smaller TED scores. The reason is that in these cases, the models tend to remember the exact words in the exemplar sentences instead of learning their POS tag information and sentence syntaxes, and thus causes the output captions just copy the words from the exemplar sentences. If we increase the replacement probability to higher value (\textit{e.g.}, 0.9), the TED score increases and the COS score decreases a little compared to those of Ours-0.7, while these two settings are comparable to each other in perplexity. Considering the overall better performance of Ours-0.7 in balancing the three metrics, setting replacement probability as 0.7 is suitable for our proposed model.

\section{Conclusion}

In this paper, we proposed a novel syntax customized video captioning task by imitating different exemplar sentences to strengthen the diversity and expressiveness of video captioning. To solve this task, a hierarchical sentence syntax encoder was proposed to capture both the local subword features and global sentence syntaxes of the exemplar sentence, based on which a two-layer syntax conditioned caption decoder was devised to generate the syntax customized caption expressing video semantic meanings. Comprehensive experiments verify that the generated captions by our model can vividly describe video contents while complying with different exemplar sentence syntaxes, thus indicating our contributions to enrich the video captioning diversity.

%



\ifCLASSOPTIONcompsoc
  \section*{Acknowledgments}
\else
  \section*{Acknowledgment}
\fi
This work was supported by the National Key Research and Development Program of China (No. 2020AAA0106300, 2018AAA0102000), National Natural Science Foundation of China Key Project (No. 62050110) and Shenzhen Nanshan District Ling-Hang Team Grant under No.LHTD20170005.

\ifCLASSOPTIONcaptionsoff
  \newpage
\fi

\bibliographystyle{IEEEtran}
\normalem
\bibliography{pami}

\begin{thebibliography}{10}
\providecommand{\url}[1]{#1}
\csname url@samestyle\endcsname
\providecommand{\newblock}{\relax}
\providecommand{\bibinfo}[2]{#2}
\providecommand{\BIBentrySTDinterwordspacing}{\spaceskip=0pt\relax}
\providecommand{\BIBentryALTinterwordstretchfactor}{4}
\providecommand{\BIBentryALTinterwordspacing}{\spaceskip=\fontdimen2\font plus
\BIBentryALTinterwordstretchfactor\fontdimen3\font minus
  \fontdimen4\font\relax}
\providecommand{\BIBforeignlanguage}[2]{{%
\expandafter\ifx\csname l@#1\endcsname\relax
\typeout{** WARNING: IEEEtran.bst: No hyphenation pattern has been}%
\typeout{** loaded for the language `#1'. Using the pattern for}%
\typeout{** the default language instead.}%
\else
\language=\csname l@#1\endcsname
\fi
#2}}
\providecommand{\BIBdecl}{\relax}
\BIBdecl

\bibitem{chen2019deep}
S.~Chen, T.~Yao, and Y.-G. Jiang, ``Deep learning for video captioning: a
  review,'' in \emph{Proceedings of the 28th International Joint Conference on
  Artificial Intelligence. AAAI Press}, 2019.

\bibitem{rohrbach2013translating}
M.~Rohrbach, W.~Qiu, I.~Titov, S.~Thater, M.~Pinkal, and B.~Schiele,
  ``Translating video content to natural language descriptions,'' in
  \emph{Proceedings of the IEEE International Conference on Computer Vision},
  2013, pp. 433--440.

\bibitem{venugopalan2015sequence}
S.~Venugopalan, M.~Rohrbach, J.~Donahue, R.~Mooney, T.~Darrell, and K.~Saenko,
  ``Sequence to sequence-video to text,'' in \emph{Proceedings of the IEEE
  international conference on computer vision}, 2015, pp. 4534--4542.

\bibitem{sutskever2014sequence}
I.~Sutskever, O.~Vinyals, and Q.~V. Le, ``Sequence to sequence learning with
  neural networks,'' in \emph{Advances in neural information processing
  systems}, 2014, pp. 3104--3112.

\bibitem{venugopalan2014translating}
S.~Venugopalan, H.~Xu, J.~Donahue, M.~Rohrbach, R.~Mooney, and K.~Saenko,
  ``Translating videos to natural language using deep recurrent neural
  networks,'' \emph{arXiv preprint arXiv:1412.4729}, 2014.

\bibitem{baraldi2017hierarchical}
L.~Baraldi, C.~Grana, and R.~Cucchiara, ``Hierarchical boundary-aware neural
  encoder for video captioning,'' in \emph{Proceedings of the IEEE Conference
  on Computer Vision and Pattern Recognition}, 2017, pp. 1657--1666.

\bibitem{pan2016jointly}
Y.~Pan, T.~Mei, T.~Yao, H.~Li, and Y.~Rui, ``Jointly modeling embedding and
  translation to bridge video and language,'' in \emph{Proceedings of the IEEE
  conference on computer vision and pattern recognition}, 2016, pp. 4594--4602.

\bibitem{wang2018reconstruction}
B.~Wang, L.~Ma, W.~Zhang, and W.~Liu, ``Reconstruction network for video
  captioning,'' in \emph{Proceedings of the IEEE Conference on Computer Vision
  and Pattern Recognition}, 2018, pp. 7622--7631.

\bibitem{xu2017learning}
J.~Xu, T.~Yao, Y.~Zhang, and T.~Mei, ``Learning multimodal attention lstm
  networks for video captioning,'' in \emph{Proceedings of the 25th ACM
  international conference on Multimedia}.\hskip 1em plus 0.5em minus
  0.4em\relax ACM, 2017, pp. 537--545.

\bibitem{yao2015describing}
L.~Yao, A.~Torabi, K.~Cho, N.~Ballas, C.~Pal, H.~Larochelle, and A.~Courville,
  ``Describing videos by exploiting temporal structure,'' in \emph{Proceedings
  of the IEEE international conference on computer vision}, 2015, pp.
  4507--4515.

\bibitem{dai2017towards}
B.~Dai, S.~Fidler, R.~Urtasun, and D.~Lin, ``Towards diverse and natural image
  descriptions via a conditional {GAN},'' in \emph{ICCV}, 2017.

\bibitem{xu2016msr}
J.~Xu, T.~Mei, T.~Yao, and Y.~Rui, ``Msr-vtt: A large video description dataset
  for bridging video and language,'' in \emph{Proceedings of the IEEE
  conference on computer vision and pattern recognition}, 2016, pp. 5288--5296.

\bibitem{guadarrama2013youtube2text}
S.~Guadarrama, N.~Krishnamoorthy, G.~Malkarnenkar, S.~Venugopalan, R.~Mooney,
  T.~Darrell, and K.~Saenko, ``Youtube2text: Recognizing and describing
  arbitrary activities using semantic hierarchies and zero-shot recognition,''
  in \emph{Proceedings of the IEEE international conference on computer
  vision}, 2013, pp. 2712--2719.

\bibitem{rohrbach2014coherent}
A.~Rohrbach, M.~Rohrbach, W.~Qiu, A.~Friedrich, M.~Pinkal, and B.~Schiele,
  ``Coherent multi-sentence video description with variable level of detail,''
  in \emph{German conference on pattern recognition}.\hskip 1em plus 0.5em
  minus 0.4em\relax Springer, 2014, pp. 184--195.

\bibitem{xu2015jointly}
R.~Xu, C.~Xiong, W.~Chen, and J.~J. Corso, ``Jointly modeling deep video and
  compositional text to bridge vision and language in a unified framework,'' in
  \emph{Proceedings of the Twenty-Ninth AAAI Conference on Artificial
  Intelligence}.\hskip 1em plus 0.5em minus 0.4em\relax AAAI Press, 2015, pp.
  2346--2352.

\bibitem{pasunuru2017reinforced}
R.~Pasunuru and M.~Bansal, ``Reinforced video captioning with entailment
  rewards,'' \emph{arXiv preprint arXiv:1708.02300}, 2017.

\bibitem{fu2018style}
Z.~Fu, X.~Tan, N.~Peng, D.~Zhao, and R.~Yan, ``Style transfer in text:
  Exploration and evaluation,'' in \emph{Proceedings of the AAAI Conference on
  Artificial Intelligence}, vol.~32, no.~1, 2018.

\bibitem{jin2020deep}
D.~Jin, Z.~Jin, Z.~Hu, O.~Vechtomova, and R.~Mihalcea, ``Deep learning for text
  style transfer: A survey,'' \emph{arXiv preprint arXiv:2011.00416}, 2020.

\bibitem{wu2020mask}
C.~Wu, X.~Chen, and X.~Li, ``Mask transformer: Unpaired text style transfer
  based on masked language,'' \emph{Applied Sciences}, vol.~10, no.~18, p.
  6196, 2020.

\bibitem{dos2018fighting}
C.~dos Santos, I.~Melnyk, and I.~Padhi, ``Fighting offensive language on social
  media with unsupervised text style transfer,'' in \emph{Proceedings of the
  56th Annual Meeting of the Association for Computational Linguistics (Volume
  2: Short Papers)}, 2018, pp. 189--194.

\bibitem{cheng2020contextual}
Y.~Cheng, Z.~Gan, Y.~Zhang, O.~Elachqar, D.~Li, and J.~Liu, ``Contextual text
  style transfer,'' in \emph{Proceedings of the 2020 Conference on Empirical
  Methods in Natural Language Processing: Findings}, 2020, pp. 2915--2924.

\bibitem{chen2020say}
S.~Chen, Q.~Jin, P.~Wang, and Q.~Wu, ``Say as you wish: Fine-grained control of
  image caption generation with abstract scene graphs,'' \emph{arXiv preprint
  arXiv:2003.00387}, 2020.

\bibitem{deshpande2019fast}
A.~Deshpande, J.~Aneja, L.~Wang, A.~G. Schwing, and D.~Forsyth, ``Fast, diverse
  and accurate image captioning guided by part-of-speech,'' in
  \emph{Proceedings of the IEEE Conference on Computer Vision and Pattern
  Recognition}, 2019, pp. 10\,695--10\,704.

\bibitem{duan2018weakly}
X.~Duan, W.~Huang, C.~Gan, J.~Wang, W.~Zhu, and J.~Huang, ``Weakly supervised
  dense event captioning in videos,'' in \emph{Advances in Neural Information
  Processing Systems}, 2018, pp. 3059--3069.

\bibitem{Gan2017StyleNet}
C.~Gan, Z.~Gan, X.~He, J.~Gao, and L.~Deng, ``Stylenet: Generating attractive
  visual captions with styles,'' in \emph{Proceedings of the IEEE Conference on
  Computer Vision and Pattern Recognition}, 2017, pp. 3137--3146.

\bibitem{Mathews2018SemStyle}
A.~Mathews, L.~Xie, and X.~He, ``Semstyle: Learning to generate stylised image
  captions using unaligned text,'' in \emph{Proceedings of the IEEE Conference
  on Computer Vision and Pattern Recognition}, 2018, pp. 8591--8600.

\bibitem{wang2019controllable}
B.~Wang, L.~Ma, W.~Zhang, W.~Jiang, J.~Wang, and W.~Liu, ``Controllable video
  captioning with pos sequence guidance based on gated fusion network,''
  \emph{arXiv preprint arXiv:1908.10072}, 2019.

\bibitem{xiao2019diverse}
H.~Xiao and J.~Shi, ``Diverse video captioning through latent variable
  expansion with conditional gan,'' \emph{arXiv preprint arXiv:1910.12019},
  2019.

\bibitem{YouImage}
Q.~You, H.~Jin, and J.~Luo, ``Image captioning at will: A versatile scheme for
  effectively injecting sentiments into image descriptions,'' \emph{arXiv
  preprint arXiv:1801.10121}, 2018.

\bibitem{hall2014less}
D.~Hall, G.~Durrett, and D.~Klein, ``Less grammar, more features,'' in
  \emph{Proceedings of the 52nd Annual Meeting of the Association for
  Computational Linguistics (Volume 1: Long Papers)}, 2014, pp. 228--237.

\bibitem{manning2014stanford}
C.~D. Manning, M.~Surdeanu, J.~Bauer, J.~R. Finkel, S.~Bethard, and
  D.~McClosky, ``The stanford corenlp natural language processing toolkit,'' in
  \emph{Proceedings of 52nd annual meeting of the association for computational
  linguistics: system demonstrations}, 2014, pp. 55--60.

\bibitem{Manning2014The}
C.~Manning, M.~Surdeanu, J.~Bauer, J.~Finkel, S.~Bethard, and D.~McClosky,
  ``The stanford corenlp natural language processing toolkit,'' in
  \emph{Proceedings of 52nd annual meeting of the association for computational
  linguistics: system demonstrations}, 2014, pp. 55--60.

\bibitem{pennington2014glove}
J.~Pennington, R.~Socher, and C.~Manning, ``Glove: Global vectors for word
  representation,'' in \emph{Proceedings of the 2014 conference on empirical
  methods in natural language processing (EMNLP)}, 2014, pp. 1532--1543.

\bibitem{krishna2017dense}
R.~Krishna, K.~Hata, F.~Ren, L.~Fei-Fei, and J.~Carlos~Niebles,
  ``Dense-captioning events in videos,'' in \emph{Proceedings of the IEEE
  international conference on computer vision}, 2017, pp. 706--715.

\bibitem{Feng2018Unsupervised}
Y.~Feng, L.~Ma, W.~Liu, and J.~Luo, ``Unsupervised image captioning,'' in
  \emph{Proceedings of the IEEE Conference on Computer Vision and Pattern
  Recognition}, 2019.

\bibitem{zhang1989simple}
K.~Zhang and D.~Shasha, ``Simple fast algorithms for the editing distance
  between trees and related problems,'' \emph{SIAM journal on computing},
  vol.~18, no.~6, pp. 1245--1262, 1989.

\bibitem{Wolf2019HuggingFacesTS}
T.~Wolf, L.~Debut, V.~Sanh, J.~Chaumond, C.~Delangue, A.~Moi, P.~Cistac,
  T.~Rault, R.~Louf, M.~Funtowicz, and J.~Brew, ``Huggingface's transformers:
  State-of-the-art natural language processing,'' \emph{ArXiv}, vol.
  abs/1910.03771, 2019.

\bibitem{Szegedy2016Inception}
C.~Szegedy, S.~Ioffe, V.~Vanhoucke, and A.~A. Alemi, ``Inception-v4,
  inception-resnet and the impact of residual connections on learning,'' in
  \emph{Thirty-First AAAI Conference on Artificial Intelligence}, 2017.

\bibitem{Russakovsky2015ImageNet}
O.~Russakovsky, J.~Deng, H.~Su, J.~Krause, S.~Satheesh, S.~Ma, Z.~Huang,
  A.~Karpathy, A.~Khosla, and M.~Bernstein, ``Imagenet large scale visual
  recognition challenge,'' \emph{International Journal of Computer Vision},
  vol. 115, no.~3, pp. 211--252, 2015.

\bibitem{KingmaAdam}
D.~P. Kingma and J.~Ba, ``Adam: A method for stochastic optimization,''
  \emph{arXiv preprint arXiv:1412.6980}, 2014.

\bibitem{paszke2019pytorch}
A.~Paszke, S.~Gross, F.~Massa, A.~Lerer, J.~Bradbury, G.~Chanan, T.~Killeen,
  Z.~Lin, N.~Gimelshein, L.~Antiga \emph{et~al.}, ``Pytorch: An imperative
  style, high-performance deep learning library,'' in \emph{Advances in Neural
  Information Processing Systems}, 2019, pp. 8024--8035.

\bibitem{Fang_2015_CVPR}
H.~Fang, S.~Gupta, F.~Iandola, R.~K. Srivastava, L.~Deng, P.~Dollar, J.~Gao,
  X.~He, M.~Mitchell, J.~C. Platt, C.~Lawrence~Zitnick, and G.~Zweig, ``From
  captions to visual concepts and back,'' in \emph{The IEEE Conference on
  Computer Vision and Pattern Recognition}, June 2015.

\bibitem{wang2019describing}
Q.~Wang and A.~B. Chan, ``Describing like humans: on diversity in image
  captioning,'' in \emph{Proceedings of the IEEE Conference on Computer Vision
  and Pattern Recognition}, 2019, pp. 4195--4203.

\end{thebibliography}

\begin{IEEEbiography}[{\includegraphics[width=1in,height=1.25in,clip,keepaspectratio]{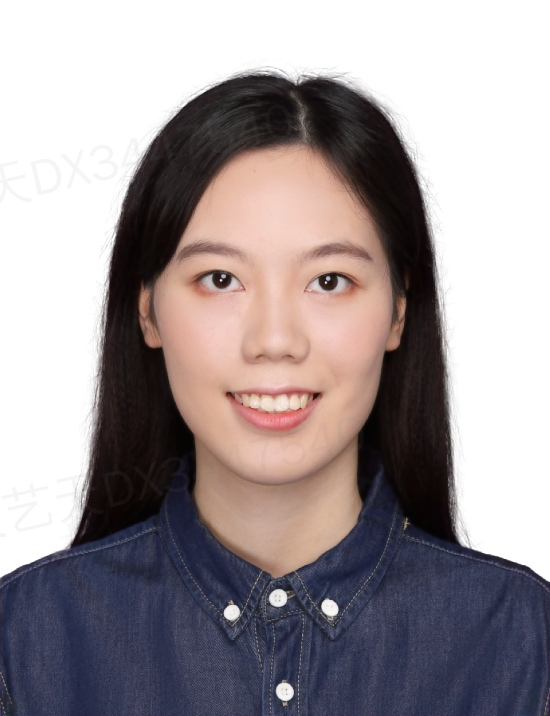}}]{Yitian Yuan} received her Ph.D. degree in computer science and technology from Tsinghua University in 2021. She got her B.E. degree in computer science from Beijing Jiaotong University in 2016. Her main research interests include multimedia analysis, computer vision and deep learning. She is currently a researcher in Meituan, Beijing, China.
\end{IEEEbiography}

\begin{IEEEbiography}[{\includegraphics[width=1in,height=1.25in,clip,keepaspectratio]{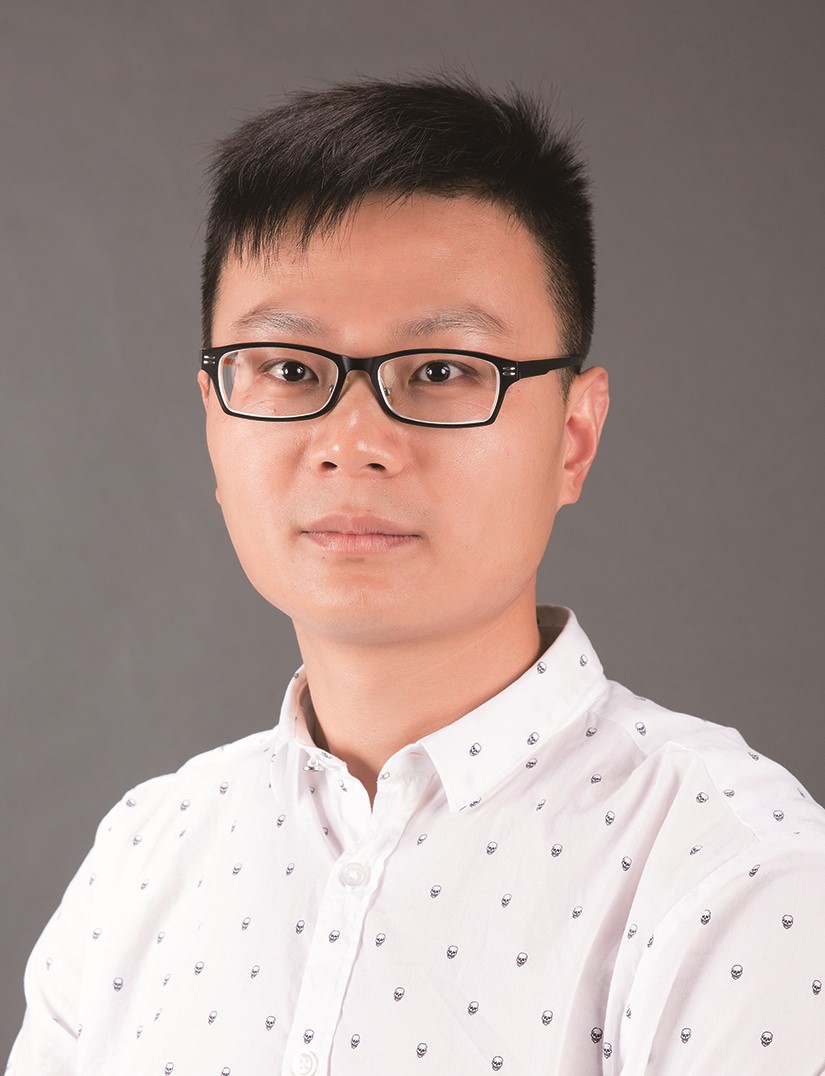}}]{Lin Ma} received the Ph.D. degree from the Department of Electronic Engineering, The Chinese University of Hong Kong, in 2013, the B.E. and M.E. degrees in computer science from the Harbin Institute of Technology, Harbin, China, in 2006 and 2008, respectively. He is now a Researcher with Meituan, Beijing, China.  Previously, he was a Principal Researcher with Tencent AI Laboratory, Shenzhen, China from Sept. 2016 to Jun. 2020. He was a Researcher with the Huawei Noah’Ark Laboratory, Hong Kong, from 2013 to 2016.  His current research interests lie in the areas of computer vision, multimodal deep learning, specifically for image and language, image/video understanding, and quality assessment.

Dr. Ma received the Best Paper Award from the Pacific-Rim Conference on Multimedia in 2008. He was a recipient of the Microsoft Research Asia Fellowship in 2011. He was a finalist in HKIS Young Scientist Award in engineering science in 2012.
\end{IEEEbiography}

\begin{IEEEbiography}[{\includegraphics[width=1in,height=1.25in,clip,keepaspectratio]{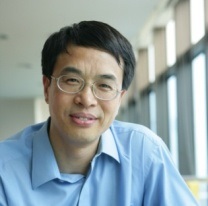}}]{Wenwu Zhu} is currently a Professor of Computer Science Department of Tsinghua University and Vice Dean of National Research Center on Information Science and Technology, Prior to his current post, he was a Senior Researcher and Research Manager at Microsoft Research Asia. He was the Chief Scientist and Director at Intel Research China from 2004 to 2008. He worked at Bell Labs New Jersey as a Member of Technical Staff during 1996-1999. 
He served as the Editor-in-Chief for the IEEE Transactions on Multimedia (T-MM) from January 1, 2017 to December 31, 2019. He has been serving as the chair of the steering committee for IEEE T-MM and Vice EiC for IEEE Transactions on Circuits and Systems for Video Technology (TCSVT) since January 1, 2020. His current research interests are in the areas of multimedia big data and intelligence, and multimedia networking including edge computing. He received nine Best Paper Awards. He is an IEEE Fellow, AAAS Fellow, SPIE Fellow, and a member of Academia Europaea.

\end{IEEEbiography}

\end{document}